%% file: main.tex
\documentclass[11pt,a4paper]{article}
\usepackage[hyperref]{emnlp2018}
\usepackage{graphicx}
\usepackage{times}
\usepackage{latexsym}
\usepackage{gb4e} 
\usepackage{multirow} 
\usepackage{url}
\usepackage[small,bf]{caption}
\usepackage{booktabs}
\usepackage{adjustbox}
\usepackage{amsmath}

\newcommand{\yg}[1]{\textcolor{red}{YG: #1}}
\newcommand{\sr}[1]{\textcolor{blue}{SR: #1}}
\newcommand{\ft}[1]{\textcolor{green}{FMT: #1}}

\renewcommand{\yg}[1]{}
\renewcommand{\sr}[1]{}
\renewcommand{\ft}[1]{}

\noautomath

\aclfinalcopy 

\title{Can LSTM Learn to Capture Agreement? The Case of Basque}

\author{Shauli Ravfogel$^1$ \and Francis M. Tyers$^{2,3}$ \and Yoav Goldberg$^{1,4}$ \\
$^1$ Computer Science Department, Bar Ilan University \\
$^2$ School of Linguistics, Higher School of Economics \\
$^3$ Department of Linguistics, Indiana University \\
$^4$ Allen Institute for Artificial Intelligence \\
  {\tt \{shauli.ravfogel, yoav.goldberg\}@gmail.com, ftyers@prompsit.com}
  }
\date{}

\begin{document}
\maketitle
\begin{abstract}
Sequential neural networks models are powerful tools in a variety of Natural Language Processing (NLP) tasks. The sequential nature of these models raises the questions: to what extent can these models implicitly learn hierarchical structures typical to human language, and what kind of grammatical phenomena can they acquire?

We focus on the task of agreement prediction in Basque, as a case study for a task that requires implicit understanding of sentence structure and the acquisition of a complex but consistent morphological system. Analyzing experimental results from two syntactic prediction tasks -- verb number prediction and suffix recovery -- we find that sequential models perform worse on agreement prediction in Basque than one might expect on the basis of a previous agreement prediction work in English. Tentative findings based on diagnostic classifiers suggest the network makes use of local heuristics as a proxy for the hierarchical structure of the sentence. We propose the Basque agreement prediction task as challenging benchmark for models that attempt to learn regularities in human language.

\end{abstract}

\section{Introduction}



In recent years, recurrent neural network (RNN) models have emerged as a
powerful architecture for a variety of NLP tasks \cite{goldberg_book}. In
particular, gated versions, such as Long Short-Term Networks (LSTMs) \cite{lstm}
and Gated Recurrent Units (GRU) \cite{cho2014learning,chung2014empirical} achieve state-of-the-art results in tasks such as language modeling, parsing, and machine translation.

RNNs were shown to be able to capture long-term dependencies and statistical regularities in input sequences \cite{karpathy-regularities, linzen2016assessing, shi-translation-syntax, jurafsky-lm, baroni2018LMagreement}.\yg{need to cite a few more here, Belvins is neither unique not a prototypical example. The least would be to add Linzen et al. But there are many others, need to add at least a few.} An adequate evaluation of the ability of RNNs to capture syntactic structure requires a use of established benchmarks. A common approach is the use of an annotated corpus to learn an explicit syntax-oriented task, such as parsing or shallow parsing \cite{dyer-parsing,goldberg-parsing, manning-parsing} \yg{cite parsing works. For example from Chris Dyer et al + Kipperwasser and Goldberg + Dozat and Mannning}. While such an approach does evaluate the ability of the model to learn syntax, it has several drawbacks. First, the annotation process relies on human experts and is thus demanding in term of resources. Second, by its very nature, training a model on such a corpus evaluates it on a human-dictated notion of grammatical structure, and is tightly coupled to a linguistic theory. Lastly, the supervised training process on such a corpus provides the network with explicit grammatical labels (e.g. a parse tree). While this is sometimes desirable, in some instances we would like to evaluate the ability of the model to implicitly acquire hierarchical representations.
 
Alternatively, one can train language model (LM) \cite{graves-lm, Jozefowicz-lm, melis-lm, yogatama2018memory} to model the probability distribution of a language, and use common measures for quality such as perplexity as an indication of the model's ability to capture regularities in language.\yg{pretty sure there are works on this line, cite? or at least recent LM works.} While this approach does not suffer from the above discussed drawbacks, it conflates syntactical capacity with other factors such as world knowledge and frequency of lexical items. Furthermore, the LM task does not provide one clear answer: one cannot be ``right'' or ``wrong'' in language modeling, only softly worse or better than other systems. 

 A different approach is testing the model on a grammatical task that does not require an extensive grammatical annotation, but is yet indicative of syntax comprehension. Specifically, previous works \cite{linzen2016assessing, bernardy2017agreement, baroni2018LMagreement} used the task of predicting agreement, which requires detecting hierarchal relations between sentence constituents. Labeled data for such a task requires only the collection of sentences that exhibit agreement from an unannotated corpora. However, those works have focused on relatively small set of languages: several Indo-European languages and a Semitic language (Hebrew).\yg{this is not accurate, baroni et al uses more than subject-verb, rephrase}\sr{Changed, alternatively it can be "a small subset of languages" without the rest of the sentence} As we show, drawing conclusions on the model's abilities from a relatively small subset of languages can be misleading. 
 
In this work, we test agreement prediction in a substantially different language, Basque, which is a language with ergative--absolutive alignment, rich morphology, relatively free word order, and polypersonal agreement (see Section \ref{sec:basque}). We propose two tasks, verb-number prediction (Section \ref{sec:predict-verb}) and suffix prediction (Section \ref{sec:suffix-pred}), and show that agreement prediction in Basque is indeed harder for RNNs. We thus propose Basque agreement as a challenging benchmark for the ability of models to capture regularities in human language. \yg{say that we'll release the agreement prediction dataset?}

\section{Background and Previous Work}
\yg{can you add also mention/description of the followup works? in particular the one from Baroni et al,
https://arxiv.org/pdf/1803.11138.pdf
but also:
https://arxiv.org/pdf/1802.09091
https://arxiv.org/pdf/1803.03585
https://arxiv.org/pdf/1805.12471
}
\yg{Instead of adding these refs, you can change the title to "Background: agreement prediction" and describe only Linzen's work. But then you need to shorten this part substantially, I propose removing the last paragraph, and maybe moving its end (where you describe basque) to the end of the intro (where we briefly describe basque.}

To shed light on the question of hierarchical structure learning, a previous
work on English \cite{linzen2016assessing} has focused on subject-verb agreement: The form of third-person
present-tense verbs in English is dependent upon the number of their subject
(``They walk'' vs. ``She walks''). Agreement prediction is an interesting case study for implicit learning of the tree structure of the input, as once the arguments of each present-tense verb in the sentence are found and their grammatical relation to the verb is established, predicting the verb form is straightforward.

\citet{linzen2016assessing} tested different variants of the agreement prediction task: categorical prediction of the verb form based on the left context; grammatical assessment of the validity of the agreement present in a given sentence; and language modeling. Since in many cases the verb form can be predicted according to number of the preceding noun, they focused on agreement attractors: sentences in which the preceding nouns have the opposite number of the grammatical subject. Their model achieved very good overall performance in the first two tasks of number prediction and grammatical judgment, while in the third task of language modeling, weak supervision did not suffice to learn structural dependencies. With regard to the presence of agreement attractors, they have shown the performance decays with their number, to the point of worse-than-random accuracy in the presence of 4 attractors; this suggests the network relies, at least to a certain degree, on local cues. \citet{bernardy2017agreement} evaluated agreement prediction on a larger dataset, and  argued that a large vocabulary aids the learning of structural patterns. \citet{baroni2018LMagreement} focused on the ability of LM's to capture agreement as a marker of syntactic ability, and used nonsensical sentences to control for semantic clues. They have shown positive results in four languages, as well as some similarities between their models' performance and human judgment of grammaticality.


\section{Properties of the Basque Language}
\label{sec:basque}
Basque agreement patterns are ostensibly more complex and very different from those of
English. In particular, nouns inflect for case, and the verb agrees with all of
its core arguments. How well can a RNN learn such agreement patterns?

We first outline key properties of Basque relevant to this work. We have used the following
two grammars written in English for reference \cite{laka96, rijk07}. 

\paragraph{Morphological marking of case and number on NPs}
The grammatical role of noun phrases is explicitly marked by nuclear case suffixes that 
attach after the determiner in a noun phrase --- this is typically the last element in the phrase.




The nuclear cases are the \emph{ergative} ({\sc erg}), the \emph{absolutive} ({\sc abs}) and the
\emph{dative} ({\sc dat}).\footnote{Additional cases encode different aspects of the role of
the noun phrase in the sentence. For example, local cases indicate aspects such
as destination and place of occurrence, possessive/genitive cases indicate possession,
causal cases indicate causation, etc. In this work we focus only on the three mentioned.} In addition
to case, the same suffixes also encode for number (singular or plural) as seen
in Table \ref{tbl:case-system}. 

\begin{table}[t]
\input{table1.tex}
\end{table}

\paragraph{Ergative-absolutive case system}
Unlike English and most other Indo-European languages that have
\emph{nominative--accusative} morphosyntactic alignment in which the single argument of intransitive
verbs and the agent of transitive verbs behave similarly to each other
(``subjects'') but differently from the object of transitive verbs, Basque has
\emph{ergative--absolutive} alignment. 
This means that the ``subject'' of an intransitive verb and
the ``object'' of a transitive verbs behave similarly to each other and receive
the absolutive case, while the ``subject'' of a transitive verb receives the ergative case.
To illustrate the difference, while in English we say ``she sleeps'' and ``she sees them''
(treating \emph{she} the same in both sentences), in an
imaginary ergative-absolutive version of English we would say ``she sleeps'' and
``her sees they'', inflecting ``she'' and ``they'' similarly (the absolutive), and
different from ``her'' (the ergative).

\paragraph{Examples}
The following sentence (\ref{ex:eman}) demonstrates the use of case suffixes to encode grammatical function.


\begin{exe}
		\ex 
\gll        \emph{Kutxazain-ek}  \emph{bezeroa-ri} \emph{liburu-ak} \emph{eman dizkiote}\\
 cashier-{\sc pl.erg} customer-{\sc sg.dat} book-{\sc pl.abs} {gave they-them-to-her/him} \\
        \trans The cashiers gave the books to the customer.
  \label{ex:eman}
\end{exe}

In (\ref{ex:eman}), the verb \emph{eman} `give' is transitive, the ergative corresponds to English 
grammatical subject and the absolutive corresponds to English grammatical object. However, 
Basque is absolutive--ergative, namely, the subject of an intransitive verb is marked for 
case like the object of a transitive verb, and differently from the subject of a transitive verb (\ref{ex:itr-daude}).
 
\begin{exe}
		\ex 
        \gll \emph{Kutxazain-ak} \emph{hemen} \emph{daude} \\
        cashier-{\sc pl-abs} here {they are-{\sc pl.abs3}} \\
        \trans The cashiers are here.
        \label{ex:itr-daude}
\end{exe}

\yg{these examples are not very clear I think, why move from the cashier and the
books and he customer to ``the men''? and not just inflect cashiers again
differently? or have the men already in the first example?}
\sr{I changed to "people"/"person" both in this example and in the above example. the first example intends to present all 3 cases, while the others focus on the ambiguity / word order.}

Since the verb \emph{daude} `are' is intransitive, the word \emph{kutxazain-} `cashier' accepts the plural 
absolutive suffix \emph{-ak}, and not the plural ergative suffix \emph{-ek}.

Interestingly, Basque exhibits case syncretism, namely, nuclear case suffixes are ambiguous: the suffix \emph{-ak} marks both plural absolutive and singular ergative. Compare
Example~(\ref{ex:per-sees-trees}) with Example~(\ref{ex:tree-sees-people}).

\begin{exe}
		\ex 
        \gll \emph{Pertson-ak} \emph{zuhaitz-ak} \emph{ikusten ditu} \\
        person-{\sc sg.erg} tree-{\sc pl.abs} he/she-sees-them \\
        \trans The person sees the trees.
        \label{ex:per-sees-trees}
\end{exe}

\begin{exe}
		\ex 
        \gll \emph{Zuhaitz-ak} \emph{pertson-ak} \emph{ikusten ditu} \\
        tree-{\sc sg.erg} person-{\sc pl.abs} {seeing it-is-them} \\
        \trans The tree sees the people.
        \label{ex:tree-sees-people}
\end{exe}

\paragraph{Word-order and Polypersonal Agreement}
Basque is often said to have a SOV word order, although the rules governing word
order are rather complex, and word order is dependent on the focus and topic of
the sentence.  While the case marking system handles most of the word-order
variation, the ambiguity between the single ergative and plural
absolutive --- which are both marked with \emph{-ak} --- results in sentence-level ambiguity. 
For instance, Example~(\ref{ex:per-sees-trees}) can also be interpreted as ``it is the
tree [{\sc sg}] that sees the people [{\sc pl}]'' (with a focus on ``the tree'').
Disambiguation in such cases depends on context and world knowledge.

Unlike English verbs that only agree in number with their grammatical subject, Basque
verbs agree in number \emph{with all} their nuclear arguments: the ergative, the absolutive and the
dative (roughly corresponding to the subject, the object and the indirect
object).\footnote{Note that some arguments, in particular proper-nouns,
are not marked for number. Other arguments, in particular the ergative, can be
omitted and not spelled out. The verb form still needs to mark the
correct number for these arguments.}
Verbs are formed in two ways: \emph{aditz trinkoak} `synthetic verbs' --- such as \emph{jakin} `to know' --- are conjugated according to the
aspect, tense and agreement patterns, e.g. \emph{dakigu} `We know it' and \emph{genekien} `We knew it'. There are only about two dozen such verbs;
all other verbs are composed of a non-finite stem, indicating the tense or aspect, and an auxiliary verb, that is conjugated according to the number of its
arguments --- such as \emph{ikusi} `to see' --- e.g. \emph{ikusten dugu} `We see it' and \emph{ikusi genuen} `We saw it'. There 
are several auxiliary verbs,
including \emph{izan} `to be' and \emph{ukan} `to have'. The form of an auxiliary verb used 
in a sentence also is also dependent on the transitivity of the verb, with \emph{izan} being
the intransitive auxiliary and \emph{ukan} being the transitive auxiliary.

\paragraph{To summarize}
Noun phrases are marked for case (ergative, absolutive or dative) and number
(singular or plural), and appear in relatively-free word
order relative to the verb to which they are arguments.
The verbs (or their auxiliaries) inflect for tense,
time and number-agreement, and agree with all their arguments on number.
Case syncretism results in ambiguity between the singular ergative and the plural absolutive
suffixes.

\section{Learning Basque Agreement}

To assess the ability of RNNs to learn Basque agreement we perform two sets of experiments.
In the first set (Section \ref{sec:predict-verb}), we focus on the ability to learn to predict the
number inflections of verbs, namely, the number of each of their arguments, where
the model reads the sentence, with one of the verbs randomly replaced with a
$\langle \textnormal{verb}  \rangle$ token.  This is analogous to the agreement
task explored in previous work on English \cite{linzen2016assessing} and other
languages \cite{baroni2018LMagreement}, but in an arguably more challenging settings,
as the Basque task requires the model: (a) to identify all the verb's arguments; (b) to
learn the ergative--absolutive distinction; and (c) to cope with a relatively
free word order and a rich morphological inflection system. As we show, the task
is indeed substantially harder than in English, resulting in much lower
accuracies than in \citet{linzen2016assessing} while not focusing on the hard cases.

\yg{consider adapting the following paragraph as a paragraph called "Limitations" at the end of the verb prediction section.}
However, we also identify some problems with the verb number prediction task.
The presence of case suffixes presumably makes the task easier, in some
sense, than in English: the grammatical role of arguments with respect to the
verb is encoded in grammatical suffixes, potentially making it easier to capture
surface heuristics that do not require the understanding of the hierarchical
structure of the sentence. In addition, the ergative---whose number is encoded
in the verb form---is often omitted from sentences, making the task of ergative
number prediction impossible without relying on context or world knowledge.
We thus propose an alternative setup (Section \ref{sec:suffix-pred}), in which,
rather than predicting the agreement pattern of the verb, we remove all nuclear
case suffixes from words and ask the model to recover them (or predict the
absence of a suffix, for unsuffixed words). We argue that this setup is a better
one for assessing models' ability to capture Basque sentence structure and
agreement system: it requires the model to accurately identify the role of NPs
with respect to a verb in order to assign them the correct case suffix (as
marked on the verb), while
not requiring the model to make-up information that is not encoded in the
sentence.
%
%

\section{Experimental Setup}
In contrast to more explicit grammatical tasks (e.g. tagging, parsing), the data
needed for training a model on agreement prediction task does not require
annotated data and can be derived relatively easily from unannotated sentences.
We have used the text of the Basque Wikipedia.
A considerable number of the articles in Basque Wikipedia appear to be
bot-generated; we have tried to filter these from the data according to
keywords. The data consists of 1,896,371 sentences; we have used 935,730 sentences for training, 129,375 for validation and 259,215 for evaluation. We make the data publicly available\footnote{\url{http://nlp.biu.ac.il/data/basque/}}. \yg{why only 65\%? is it due to
the removal of the bot-generated sentences? if so, it is not clear.}

\sr{There is no really a reason. But note this doesn't matter in practice since the model ceases to improve long before the completion of the first epoch - it stops improving after about 600,000 sentences which is approx. a third of the data}\yg{still need to explain why you did this. Or you can just not say that the data consists of 1,896,371 sentences, and just say ``From the remaining data, we use XYZ sentences from training and ABC for evaluation.'' btw, no dev set? or is it part of the train? if part of train, then be explicit about the train/dev/test split sizes.}

We use the Apertium morphological analyzer \cite{apertium, eues} to extract the
parts-of-speech (POS) and morphological marking of all words.\footnote{We use the Apertium analyzer instead of other options as it is freely available online under a free/open-source licence covering both the lexicon and the source code.} The POS information was used to detect verbs, nouns and adjectives, but was not incorporated in the word embeddings. 

For section \ref{sec:diagnostic-treebank}, grammatical generalization, we used the Basque Universal Dependencies treebank \cite{aranzabe15} to extract human-annotated POS, case, number and dependency edge labels. We have used their train:dev:test division, resulting in 5,173 training sentences, 1,719 development sentences and 1,719 test sentences.

\paragraph{Word Representation} We represent each word with an embedding vector. To account for the rich morphology of Basque, our word embeddings combine the word identity, its lemma\footnote{Most words admit to a single interpretation by the morphological analyzer. For words that had several optional lemmas, we chose the first one, after the exclusion of colloquial or familiar verb forms, which are infrequent in Wikipedia.} as determined by the morphological analyzer, and character ngrams of lengths 1 to 5. Let $E_t$, $E_l$ and $E_{ng}$ be token, lemma and n-gram embedding matrices, and let $t_{w}$, $l_{w}$ and $\{ng_w\}$ be the word token, the lemma and the set of all n-grams of lengths 1 to 5, for a given word $w$. The final vector representation of $w$, $e_w$, is given by $e_w$ = $E_{t}[t]$ + $E_{l}[l]$ + $\sum\nolimits_{ng \in \{ng_w\}} E_{ng}[ng]$. We use embedding vectors of size 150. We recorded the 100,000 most common words, n-grams and lemmas, and used them to calculate the vector representation of words. Out-of-vocabulary words, ngrams and lemmas are replaced by a $\langle \textnormal{unk} \rangle$ token.
 
 \paragraph{Model} In previous studies, the agreement was between two
 elements, and the model was tasked with predicting a morphological property of
 the second one, based on a property encoded on the first. Thus, a
 uni-directional RNN sufficed. Here, due to a single verb having to agree with several
 arguments, while following a relatively free word order, we cannot use a
 uni-directional model. We opted instead for a bi-directional RNN.\footnote{A unidirectional LSTM baseline achieved accuracy scores of 86.6\%, 91.7\% and 98.2\% and recall values of 78.9\%, 100\% and 60.1\% for ergative, absolutive and dative verb arguments prediction, respectively.}  In all
 tasks, we use a one-layer BiLSTM network with 150 hidden units, compared with 50 units in \cite{linzen2016assessing} \footnote{Network size was chosen based on development set performance.}. 
 In the verb
 prediction task, the BiLSTM encodes the verb in the context of the entire
 sentence, and the numbers of the ergative, absolutive and datives are predicted by 3 independent multilayer perceptrons (MLPs) with a single hidden layer of size 128, that receive as an input the hidden state of the BiLSTM over the $\langle \textnormal{verb} \rangle$ token.

 In the suffix prediction task, the prediction of the case suffix is performed by a MLP of size 128, that receives as an input the hidden state of the BiLSTM over each word in the sentence.

The whole model, including the embedding, is trained end-to-end with the Adam optimizer \cite{adam}.

\section{Verb Argument Number Prediction}
\label{sec:predict-verb}

In this task, the model sees the sentence with one of the auxiliary verbs replaced by a $\langle
\textnormal{verb} \rangle$ token, and predicts the number of its ergative,
absolutive and dative. For example, in (\ref{ex:eman}) above,
the network sees the embeddings of the words in the sentence:\footnote{See:
\begin{exe}
		\ex 
\gll        \emph{Kutxazain-ek}  \emph{bezeroa-ri} \emph{liburu-ak} \emph{eman} \emph{$\langle \textnormal{verb} \rangle$} \\
	    {cashier-{\sc pl.erg}} {customer-{\sc pl.dat}} {book-{\sc sg.abs}} {give-{\sc ptcp}} {$\langle \textnormal{aux} \rangle$} \\
  \trans `The cashiers gave the books to the customers'
\end{exe}
}
\setlength\tabcolsep{1.5pt} 
\begin{center}
\begin{tabular}{ccccc}
 \emph{Kutxazain-ek} &  \emph{bezeroa-ri}  & \emph{liburu-ak}  & \emph{eman} & \emph{$\langle \textnormal{verb} \rangle$} \\
\end{tabular}\\
\end{center}
\setlength\tabcolsep{6pt} 

It is then expected to predict the number of the arguments of the missing verb,
\emph{dizkiote}: ergative:plural, dative:singular and absolutive:plural. Each argument can take one of three values, \emph{singular}, \emph{plural} or \emph{none}.
In order to succeed in this task, the model has to identify the arguments of the
omitted verb, and detect their plurality status as well as their grammatical relation to the verb. Note that as
discussed above, these relations do not overlap with the notions of ``subject''
and ``object'' in English, as the grammatical case is also dependent on the
transitivity of the verb. Since the model is exposed to the lemma of the auxiliary verb and the stem that precedes it, it can, in principle, learn dividing verbs into transitive and intransitive.\yg{does it learn to do
so? worth checking with diagnostic classifers? or simply with projections to 2d?} 
\sr{There are 3 main auxiliary verbs. It's hard to tell if the model learns the generalization of "transitive verb" or memorizes which verbs are transitive. The embeddings of the auxiliary verbs - see the plot embeddings-auxiliary.png - show it groups them according to the number of the arguments, not according to transitivity. }

\subsection{Results and Analysis}

We conducted a series of experiments, as detailed below. 
A summary of all the results in available in Table \ref{tbl:verb-summary}.

\begin{table}[t]
    \scalebox{0.8}{
\begin{tabular}{l|rrr}
\toprule
    \textbf{Condition} & \textbf{Ergative} & \textbf{Absolutive} & \textbf{Dative} \\
              & A / R    & A / R      & A / R \\
    \midrule
    Base & 87.1 / 80.0  & 93.8 / 100  & 98.0 / 54.9 \\
    \midrule
    Suffixes only & 69.0 / 40.3 & 83.7 / 100 & 97.0 / 26.0 \\
    No suffixes & 83.8 / 80.0 & 87.8 / 100 & 97.3 / 34.7 \\
    Neutralized case & 86.0 / 79.3 & 93.3 / 100 & 97.3 / 38.1 \\
    \midrule
    Single verb & 90.6 / 89.0 & 96.04 / 100 & 98.9 / 74.7 \\
    No \emph{-ak} & 90.9 / 81.1 &  96.6 / 100 & 98.6 / 67.7 \\
    Sing. verb, no \emph{-ak} & 92.6 / 83.4 & 97.2 / 100 & 99.1 / 75.4 \\
\bottomrule
\end{tabular}}
\caption{Summary of verb number prediction results for accuracy (A) and recall (R).}
\label{tbl:verb-summary}
\end{table}

\paragraph{Main results} The model achieved moderate success in this task, with accuracy of 87.1\% and
93.8\% and recall of 80.0\% and 100\%\footnote{This reflects the fact the
absolutive is almost always present.} in ergative and absolutive prediction,
respectively. Dative accuracy was 98.0\%, but the recall is low (54.9\%),
perhaps due to the relative rarity of dative nouns in the corpus (only around 3.5\% of
the sentences contain dative). These relatively low numbers are in sharp contrast to previous
results on English in which the accuracy scores on general sentences was above 99\%. While
English agreement results drop when considering hard cases where agreement
distractors or intervening constructions intervene between the verb and its
argument, in Basque the numbers are low already for the common cases.

This suggests that agreement prediction in Basque can serve as a valuable
benchmark for evaluating the syntactic abilities of sequential models such as
LSTMs in a relatively challenging grammatical environment, as well as for assessing the generality of results across language families.

\paragraph{Ablations: case suffixes vs. word forms}
The presence of nuclear case suffixes in Basque can, in principle, make the task
of agreement prediction easier, as (ambiguous) grammatical annotation is
explicit in the form of the nuclear case suffixes, that encode the type of
grammatical connection to the verb. 
How much of the relevant information is encoded in the case suffixes?
To investigate the relative importance of
these suffixes, we considered a baseline in which the model is exposed only to
the nuclear suffixes, ignoring the identities of the words and the character
n-grams (Table \ref{tbl:verb-summary}, Suffixes only). This model achieved accuracy scores of 69.0\%,  83.7\% and 97.0\% and recall values of 40.3\%, 100\% and 26\% for ergative, absolutive and dative prediction, respectively. While substantially lower than when
considering the word forms, the absolute numbers are not random,
suggesting that agreement can in large part be predicted based on the presence
of the different suffixes and their linear order in the sentence, without paying
attention to specific words. 

In a complementary setting the model is exposed to the sentence after the removal of all nuclear case suffixes (according to the morphological analyzer output). This setting (Table \ref{tbl:verb-summary}, No suffixes) yields accuracies of 83.8\%, 87.8\% and 97.3\% and recall scores of 80.0\%, 100\% and 34.7\% for ergative, absolutive and dative, respectively. Interestingly, in the last setting the model succeeds to some extent to predict the verb arguments number although the number is not marked on the arguments. This suggests the model uses cues such as the existence of certain function words that imply a number, and the forms of non-nuclear suffixes to infer the number of the arguments.
%
%

\paragraph{Importance of explicit case marking}
The verb numbers prediction task requires the model to identify the arguments, and hence
the hierarchical structure of the sentence. However, the Basque suffixes encode
not only the number but also the explicit grammatical function of the argument.
This makes the
model's task potentially easier, as it may make use of the explicit case
information as an effective heuristic instead of modeling the sentence's
syntactic structure.
To control for this, we consider a neutralized version (Table
\ref{tbl:verb-summary}, Neutralized case) in which we removed case
information and kept only the number information: suffixes were replaced by
their number, or were marked as ``ambiguous'' in case of \emph{-ak}. For example,
the word kutxazainek was replaced with kutxazain$\langle \textnormal{plural}
\rangle$, since the suffix \emph{-ek} encodes plural ergative. Interestingly, in
this settings the performance was only slightly decreased, with accuracy scores
of 86.0\%, 93.3\% and 97.3\% and recall values of 79.3\%, 100\% and 38.1\% for
ergative, absolutive and dative, respectively. These results suggest that the
network either makes little use of explicit grammatical marking in the suffixes,
or compensates for the absence of grammatical annotation using other information
present in the sentence.
\paragraph{Performance on simple sentences}
The presence of multiple verbs, along with the inherent ambiguity of the
suffix system, can both complicate the task of number prediction. To assess
the relative importance of these factors, we considered modified
training and test sets that contain only sentences with a single verb (Table
\ref{tbl:verb-summary}, Single verb). This
resulted in a significant improvement, with accuracy scores of 90.61\%, 
96.04\% and 98.9\% and recall values of 89.0\%, 100\% and 74.7\% for ergative, absolutive, and dative, respectively; note that sentences with a
single verb also tend to be shorter and simpler in their grammatical structure.
To evaluate the influence of the ambiguous suffix, we removed all sentences that
contain the ambiguous suffix \emph{-ak} from the dataset (Table
\ref{tbl:verb-summary}, No \emph{-ak}). This resulted in a more moderate improvement to accuracy values of 90.9\%, 96.6\% and 98.6\% and recall of 81.1\%, 100\% and 67.7\% for ergative, absolutive and dative. Limiting the dataset to unambiguous sentences with a single verb
(Table \ref{tbl:verb-summary}, Sing. verb, no \emph{-ak}) yields an additional improvement, with accuracies of 92.6\%, 97.2\% and 99.1\% and recall values of 83.4\%, 100\% and 75.4\% for ergative, absolutive and dative, respectively.

\section{NP Suffix Prediction}
\label{sec:suffix-pred}
The general trend in the experiments above is a significantly higher success in
absolutive number prediction, compared with ergative number prediction. This
highlights a shortcoming in the verb-number prediction task: as Basque encodes
the number of the verb arguments in the verb forms, the subject can --- and often
is --- be omitted from the sentence. Additionally, the number of proper nouns is
often not marked. These cases are common for the ergative: 55\% of the sentences marked for {\sc erg.pl3} agreement do not contain words suffixed with \emph{-ek}. \yg{is this true?} \sr{yes: 55\% of the sentences that contain a {\sc erg.pl3} agreement don't contain -ek suffixed words}\yg{added this number to the text.}
This
requires the model to predict the number of the verb based on information which
is not directly encoded in the sentence.
\ft{What do you mean by "the number of proper nouns is not marked" ?}\yg{That's my understanding, which may be wrong: number is marked on the determiner, but proper nouns do not have a determiner. Isn't that so? my source is https://www.ehu.eus/documents/2430735/0/A-brief-grammar-of-euskara.pdf}

To counter these limitations, we propose an alternative prediction task that
also takes advantage of the presence of case suffixes, while not
requiring the model to guess based on unavailable information. In this task, the network reads the input sentence with all nuclear case suffixes removed, and has to predict the suffix (or the absence of thereof) for each word in the sentence. For example, in (\ref{ex:eman}) above, the model reads (\ref{ex:lemmatised}).

\begin{exe}
		\ex  Kutxazaina bezeroa liburua eman dizkiote.
               \label{ex:lemmatised}
\end{exe}

It is then expected to predict the omitted case and determiner suffixes (\emph{-ek}, \emph{-ak}, \emph{-ari}, none,
none). We note that we remove the suffixes only from NPs, keeping the verbs
in their original forms. As the verbs encode the numbers of its argument as well
as their roles, the network is exposed to all relevant information required for
predicting the missing suffixes, assuming it can recover the sentence structure. 
In order to succeed in this task, the model should link each argument to its
verb, evaluate its grammatical relation to the verb, and choose the case suffix
accordingly. Case suffixes are appended at the end of the NP.
\ft{A better way of putting this might be that the case and determiner suffixes are appended at the end of the NP, so if you have "num noun adj" then it goes on the adj but if you have "num noun" it goes on the noun, it's basically slapped on the last component of the NP.}
As a result, suffix
recovery also requires some degree of POS tagging and NP chunking, and thus
shares some similarities with shallow parsing in languages such as English. This
suggests that the task of case suffix recovery in languages with complex case
system such as Basque can serve as a proxy task for full parsing, while
requiring a minimal amount of annotated data.\yg{For the future: can we improve
the accuracy of a basque parser by performing multi-task training on the suffix
prediction task? worth considering.}
%
%

%
%

The singular absolutive determiner suffix, \emph{-a}, also appears in the base form of some words. Therefore, for \emph{-a} suffixed words, we have used the morphological analyzer to detect whether not the \emph{-a} suffix is a part of the lemma. Consider the examples \emph{ur} `water'---\emph{ura} `the water-{\sc abs}' and \emph{uda} `summer'---\emph{uda} `the summer-{\sc abs}'. \emph{-a} suffixed words not known to the analyzer were excluded from the experiment.

\begin{table}[t]
\input{table2.tex}

\end{table}

\subsection{Results and Analysis}

The results for the suffix prediction task are presented in Table
\ref{tbl:suff-results} and Table \ref{tbl:suffix-summary}. The model achieves F1 scores of 78.2 and 83.2\% for the erg. plural \emph{-ek} and absolutive singular/ergative singular \emph{-ak} suffixes, respectively. The F1 score for the {\sc abs} singular suffix \emph{-a} is higher --- 85.5\%; This might be due to the fact this suffix is unambiguous (unlike \emph{-ak}), and the fact the absolutive is rarely omitted (unlike words suffixed with \emph{-ek}), which implies that verb forms indicating verb-absolutive singular agreement also reliably predicts the presence of a word suffixed with -a in the sentence. Similarly to the trend in the first task, the model achieved relatively low F1 scores in the prediction of dative suffixes, \emph{-ari} and \emph{-ei}: 78.8\% and 65.0\%, respectively. 

\begin{table}[t]
\begin{center}
    \scalebox{0.8}{

\begin{tabular}{l|rrrrr}
\toprule
          & \emph{-ak}   & \emph{-ek} & \emph{-a} & \emph{-ari}  & \emph{-ei}   \\ \midrule
Base      & 83.2 & 78.2 & 85.5 & 78.8& 65.0 \\
Word-only & 56.0 & 49.5 & 55.2 & 56.5 & 24.2 \\
No verb   & 72.0 & 65.4 & 78.1 & 67.5  & 47.3  \\
\bottomrule
\end{tabular}}
\end{center}
\caption{Summary of F1 scores for suffix prediction.}
\label{tbl:suffix-summary}
\end{table}

\paragraph{Importance of verb form} \sr{Isn't "Importance of verb form" more accurate?}\yg{feel free to change.}
Once the grammatical connection between verbs and their arguments is established, the nuclear suffix of each of the verb's arguments is deterministically determined by the form of the verb. As such, verb forms are expected to be of importance for suffix prediction. To assess this importance, we have evaluated the model in a setting in which the original verb forms are replaced by a $\langle \textnormal{verb}  \rangle$ token. In this setting, the model achieved F1 scores of 72.0\%, 65.4\%, 78.1\%, 67.5\%, 47.3\% and 92.0\% for \emph{-ak}, \emph{-ek}, \emph{-a}, \emph{-ari}, \emph{-ei}, and the prediction of the presence of any nuclear suffix, respectively (Table \ref{tbl:suffix-summary}, No verb). These results, that are far from random, indicate that factors such as the order of words in the sentence, the identity of the words (as certain words tend to accept certain cases irrespective of context), and the non-nuclear case suffixes (which are not omitted), all aid the task of nuclear-suffix prediction.
\yg{would be nice o have a summary table of all ablations, like I added in the prev
section, or at least compare the numbers here to the relevant numbers directly
in the text.} \sr{Added tbl:suffix-summary}

\paragraph{Word-only baseline} Some words tend to appear more frequently in
certain grammatical positions, regardless of their context. We therefore
compared the model performance with a baseline of a 1-layer MLP that predicts
the case suffix of each word based only its embedding vector. As expected, this baseline achieved lower F1 scores of 56.0\%, 49.5\%, 55.2\%, 56.5\%, 24.2\% and 69.8\% for \emph{-ak}, \emph{-ek}, \emph{-a}, \emph{-ari}, \emph{-ei}, and the prediction of the presence of any suffix, respectively (Table \ref{tbl:suffix-summary}, Words only).

%
%

\paragraph{Focusing on the harder cases}

\begin{table*}[t]
  \centering

  \resizebox{0.8\textwidth}{!}{\begin{minipage}{\textwidth}
\begin{center}

\input{table3.tex}
\end{center}
      \end{minipage}}

\end{table*}

\begin{table*}[t]
  \centering

  \resizebox{0.8\textwidth}{!}{\begin{minipage}{\textwidth}

\begin{center}
\input{table4.tex}
\end{center}
      \end{minipage}}
\end{table*}

\ft{Not sure what you mean by "their constituents" below, perhaps "verbs and other constituents in the sentence" ? -- but I think "identifying arguments" works as well}
An essential step in the process of suffix prediction is identifying the arguments of each verb. To what extent does the model rely on
local cues as a proxy for this task? A simple heuristics is relating each word
to its closest verb. We compared the model's performance on ``easier'' instances
where the closest verb is grammatically connected to the word, versus 
``harder'' instances in which the closest verb is not grammatically connected to the word.
 
This evaluation requires automatically judging the grammatical connection
between words and verbs in the input sentence. Due to the ambiguous case
suffixes, this is generally not possible in unparsed corpora. However, we focus
on several special cases of sentences containing exactly 2 verbs of specific
types, in which it is possible to unambiguously link certain words in the
sentences to certain verbs. Since these instances consists only a fraction of
the dataset\yg{what fraction? what is the new train/test sizes?}\sr{changes from instance to instance, a few thousands sentences. The numbers appear in the results doc, and the numbers for "da" are mentioned in the table. Do you think we should present the full data here?}, for this evaluation we have used a larger test set containing 50\% of the data.

%
%
%
\ft{In general I think "constituent" is better than "constituency" --- unless you have a specific definition of the latter term.}
Table \ref{tbl:da} depicts the results for sentences that contain the verb
\emph{da} `is'. The general trend, for \emph{da} and for several other verbs (not presented
here \yg{which ones?}) \sr{it appears in the text above, currently commented. There are 3 verbs: da, zen, ziren}, is higher F1 scores in the ``easier'' instances. We
note, however, that in these instances there is also larger absolute distance
between the verb and its argument, which prevents us from drawing causal conclusions.
%
%
\ft{In table2 both sides say "closest verb is incorrect" --- is that right?}
\paragraph{Diagnostic classifiers}
To overcome this difficulty and understand if the model encodes the grammatical connection between a word to its closest verb in the BiLSTM hidden state over a given word, we have trained a diagnostic classifier \cite{diagnostic1, diagnostic2} that receives as an input the hidden state of a BiLSTM over a word, and predicts whether or not the closest verb (which is unseen by the diagnostic classifier) was grammatically connected to the word.

We have compared two diagnostic classifiers: a linear model, and a 1-layer MLP. A training set was created by collecting hidden states of the BiLSTM over words, and labeling each training example according to the existence of a verb-argument connection between the word over which the state was collected and its closest verb (a binary classification task). We then compared the success rate of the diagnostic classifier on instances in which the BiLSTM correctly predicted a case suffix (Table \ref{tbl:diagnostic-results}, BiLSTM correct), versus the instances on which the BiLSTM predicted incorrectly (Table \ref{tbl:diagnostic-results}, BiLSTM wrong). 
The results, depicted in Table \ref{tbl:diagnostic-results}, demonstrate that in instances in which the model predicts a wrong case
suffix, the diagnostic classifier tends to inaccurately predict the connection
between the closest verb and the word. For example, for sentences that contain
the verb form \emph{da}, the success rate of the linear model increases from 56.2\% to 70.2\% in the instances in which the BiLSTM predicted correctly. This differential success may imply a causal relation between the inference of the closest-verb grammatical connection to the word and the success in suffix prediction. 

%
%

 

\paragraph{Grammatical generalization}
\label{sec:diagnostic-treebank}

Does training on suffix recovery induce learning of grammatical generalizations such as morphosyntactic alignment (ergative, absolutive or dative), number agreement (sg / pl) and POS? To test this question, We have collected the states of our trained model over the words in sentences from the Basque Universal Dependencies dataset. Different diagnostic classifiers were then trained to predict case, number, POS and the type of the dependency edge to the head of the word. All diagnostic classifiers are MLPs with two hidden layers of sizes 100 and 50. For each task, we trained 5 models with different initializations and report those that achieved highest development set accuracy.

For nuclear case and number prediction, we limit the dataset to words suffixed with a nuclear case. In this setting, for words on which the BiLSTM predicted correctly, the MLPs perform well, predicting the correct number with an accuracy of 95.0\% (majority classifier: 67.3\%) and the correct case with an accuracy of 93.5\% (majority: 61.7\%). Even when the dataset is limited to words suffixed with the ambiguous suffix \emph{-ak}, the MLP correctly distinguishes ergative and absolutive with 91.2\% accuracy (majority: 65.4\%). Interestingly, in a complementary setting on which the dataset is limited to words on which the BiLSTM failed in nuclear case suffix recovery, a diagnostic classifier can still be trained to achieve 74.7\% accuracy in number prediction and 69.7\% accuracy in case prediction. This indicates that to a large degree, the required information for correct prediction is encoded by the state of the model even when it predicts a wrong suffix.  

For the prediction of POS, dependency edge to the head and any case (not just nuclear cases --- 16 cases in total, including the option of an absence of case), the dataset was not limited to words suffixed with nuclear cases or to words on which the BiLSTM predicted correctly. The classifier achieves accuracies of 87.5\% In POS prediction (majority: 23.2\%), 85.7\% in the prediction of any case (majority: 64.7\%), and 69.0\% for the prediction of dependency edge to the head (majority: 19.0\%).

These results indicate that during training on suffix recovery, the model indeed learns, to some degree, the generalizations of number, alignment and POS, as well as some structural information (connection to the head in the dependency tree). These findings support our hypothesis that a success in case recovery entails the acquiring of some grammatical information.

\section{Conclusion}

In this work, we have performed of series of controlled experiments to evaluate the performance of LSTMs in agreement prediction, a task that requires implicit understanding of syntactic structure. We have focused on Basque, a language that is characterized by a very different grammar compared with the languages studied for this task so far. We have proposed two tasks for the evaluation of agreement prediction: verb number prediction and suffix recovery. 

Both tasks were found to be more challenging than agreement prediction in other languages studied so far. We have evaluated different contributing factors to that difficulty, such as the presence of ambiguous case suffixes. We have used diagnostic classifiers to test hypotheses on the inner representation the model had acquired, and found tentative evidence for the use of shallow heuristics as a proxy of hierarchical structure, as well as for the acquisition of grammatical information during case recovery training.

These results suggest that agreement prediction in Basque could be a challenging benchmark for the evaluation of the syntactic capabilities of neural sequence models. The task of case-recovery can be utilized in other languages with a case system, and provide a readily-available benchmark for the evaluation of implicit learning of syntactic structure, that does not require the creation of expert-annotated corpora. A future line of work we suggest is investigating what syntactic representations are shared between case recovery and full parsing, i.e., to what extent does a model trained on case recovery learn the parse tree of the sentence, and whether transfer learning from case-recovery would improve parsing performance. 

\section*{Acknowledgements}
We would like to thank Mikel L. Forcada and Mikel Artetxe for discussions regarding Basque grammar and the anonymous reviewers
for their helpful comments.

\bibliographystyle{acl_natbib_nourl}
\bibliography{refs}

\end{document}

%% file: table1.tex
\begin{center}
\begin{tabular}{lllll}
\toprule
\multirow{2}{*}{\textbf{Case}} & \multirow{2} {*}{\textbf{Function}}        & \multicolumn{3}{c}{\textbf{Suffix Forms}}                           \\ 
                               &                                          & \textbf{Sg} & \textbf{Pl} & \textbf{No det} \\ \midrule
Absolutive                     & {\sc S}, {\sc O} & \emph{-a}                & \emph{-ak}             & -                      \\
Ergative                       & {\sc A}                       & \emph{-ak}               & \emph{-ek}             & \emph{-(e)k}                  \\
Dative                         & {\sc IO} & \emph{-ari}              & \emph{-ei}            & \emph{-(r)i}                  \\
\bottomrule
\end{tabular}
\caption{\label{tbl:case-system} Basque case and their corresponding determined nuclear case suffixes. Note the case syncretism, resulting in structural ambiguity
   between the plural absolutive and the ergative singular. Under function, {\sc S} refers to the single argument of a prototypical
   intransitive verb, {\sc O} refers to the most patient-like argument of a prototypical transitive verb, and {\sc A} refers 
   to the most agent-like argument of a prototypical transitive verb. Subsequently {\sc IO} refers to the indirect object, often
   filling the recipient or experiencer role.}
\end{center}

%% file: table2.tex
\begin{tabular}{lrrr}
\toprule
\textbf{Suffix}                                   & \textbf{Prec } & \textbf{Rec } & \textbf{F1 } \\ \midrule
\emph{-ek} {[}ergative plural{]}                          & 82.0                    & 74.7                 & 78.2             \\
\emph{-a} {[}absolutive singular{]}                       & 88.0                    & 83.2                & 85.5             \\
\emph{-ak} {[}abs. pl / erg. sg{]}              & 83.2                    & 83.1                 & 83.2             \\
\emph{-ari} {[}dative singular{]}                         & 80.2                    & 77.5                 & 78.8             \\
\emph{-ei} {[}dative plural{]}                            & 65.5                    & 64.5                 & 65.0             \\
Any                                               & 95.1                    & 91.7                 & 93.4            \\
\bottomrule
\end{tabular}
\caption{\label{tbl:suff-results} nuclear case prediction results. }

%% file: table3.tex
\begin{tabular}{llll|lll}
\toprule
\textbf{Suffix}         & \multicolumn{3}{c|}{Closest verb is incorrect} & \multicolumn{3}{c}{Closest verb is correct}        \\ \hline
\multicolumn{1}{l|}{}   & Rec              & Prec            & F1        & Rec           & Prec          & F1                   \\ \cline{2-7} 
\multicolumn{1}{l|}{\emph{-ak}} & 70.2 (2961)    & 85.2 (2438)   & 76.9    & 80.5 (8312) & 88.2 (4954)   & 84.1 \\
\multicolumn{1}{l|}{\emph{-ek}} & 60.8 (758)     & 98.2  (469)   & 75.1    & 64.7 (1976) & 95.9 (1333) & 77.2               \\
\bottomrule
\end{tabular}

\caption{\label{tbl:da} Model performance according to closest-verb grammatical connection to the verb, for sentences that contain the verb \emph{da} `it is'. The number of sentences appears in parentheses. }
 

%% file: table4.tex
\begin{flushleft}


\begin{tabular}{llllll}
\toprule
\multirow{2}{*}{Verb form} & \multirow{2}{*}{Diagnostic classifier} & \multicolumn{3}{c}{Accuracy (\%)}                                                                                                                                     & \multirow{2}{*}{Majority (\%)} \\
                      &                                        & Total & \begin{tabular}[c]{@{}l@{}}BiLSTM wrong\end{tabular} & \begin{tabular}[c]{@{}l@{}}BiLSTM correct\end{tabular} &                                      \\ \midrule
\emph{da} `is'                   & Linear model                           & 67.7  & 56.2                                                                         & 70.2                                                                           & \multirow{2}{*}{62.4}                \\
                      & 1-layer MLP                            & 74.7  & 69.3                                                                         & 75.6                                                                           &                                      \\
\emph{zen} `was'                  & Linear model                           & 64.4  & 52.8                                                                         & 66.5                                                                           & \multirow{2}{*}{61.9}                \\
                      & 1-layer MLP                            & 74.8  & 71.5                                                                         & 75.4                                                                           &                                      \\
\emph{ziren} `were'                & Linear model                           & 67.4  & 57.8                                                                         & 70.1                                                                           & \multirow{2}{*}{59.8}                \\
                      & 1-layer MLP                            & 76.6  & 72.3                                                                         & 77.8                                                                           &                                    \\
\bottomrule 
\end{tabular}

\caption{\label{tbl:diagnostic-results}
 Diagnostic classifier accuracy in predicting whether or not the closest verb is grammatically connected to a word, according to BiLSTM suffix prediction success on that word. ``BiLSTM correct": success rate on instances in which the BiLSTM correctly predicted the case suffix. `BiLSTM wrong": success rate on instances in which the BiLSTM failed. ``Majority" signifies the success of majority-classifier. }

\end{flushleft}